\documentclass{article}
\usepackage{ijcai22}

\usepackage{times}
\usepackage{soul}
\usepackage{url}
\usepackage{bm}
\usepackage[hidelinks]{hyperref}
\usepackage{hyperref}
\usepackage[utf8]{inputenc}
\usepackage[small]{caption}
\usepackage{graphicx}
\usepackage{amsmath}
\usepackage{mathrsfs}
\usepackage{amsthm}
\usepackage{booktabs}
\usepackage{algorithm}
\usepackage{algorithmic}
\usepackage{booktabs}
\usepackage{amssymb}
\usepackage{color}
\usepackage{subfigure}
\usepackage[table,xcdraw]{xcolor}
\graphicspath{{Figure/}}

\title{Instance-aware Prompt Learning for Language Understanding and Generation}

\author{
Feihu Jin$^{1,2}$\
\and
Jinliang Lu$^{1,2}$\and
Jiajun Zhang$^{1,2}$ \footnote{Corresponding Author}\And
Chengqing Zong$^{1,2,3}$
\affiliations
$^1$National Laboratory of Pattern Recognition, Institute of Automation, CAS\\
$^2$School of Artificial Intelligence, University of Chinese Academy of Sciences\\
$^3$CAS Center for Excellence in Brain Science and Intelligence Technology
\emails
jinfeihu2020@ia.ac.cn,
\{jinliang.lu, jjzhang, cqzong\}@nlpr.ia.ac.cn,
}

\begin{document}

\maketitle

\begin{abstract}
Recently, prompt learning has become a new paradigm to utilize pre-trained language models (PLMs) and achieves promising results in downstream tasks with a negligible increase of parameters. The current usage of discrete and continuous prompts assumes that the prompt is fixed for a specific task and all samples in the task share the same prompt. However, a task may contain quite diverse samples in which some are easy and others are difficult, and diverse prompts are desirable. In this paper, we propose an instance-aware prompt learning method that learns a different prompt for each instance. Specifically, we suppose that each learnable prompt token has a different contribution to different instances, and we learn the contribution by calculating the relevance score between an instance and each prompt token. The contribution weighted prompt would be instance aware. We apply our method to both unidirectional and bidirectional PLMs on both language understanding and generation tasks. Extensive experiments demonstrate that our method obtains considerable improvements compared to strong baselines. Especially, our method achieves the state-of-the-art on the SuperGLUE few-shot learning benchmark.$\footnote{Our code is available in\ \textcolor{blue}{\url{https://github.com/jinfeihu-stan/IPL}}}$

\end{abstract}

\section{Introduction}
Prompt learning aims to design or learn appropriate prompts which can induce the capacity from pre-trained language models (PLMs) to perform specific tasks, and it becomes a new paradigm to use PLMs due to its flexibility and fewer extra parameters. There are typically two kinds of prompts, namely discrete prompt and continuous prompt. Discrete prompt such as GPT-3 \cite{Brown2020LanguageMA} uses the task instructions and task-related instances as prompt for zero-shot and few-shot learning respectively. PET/iPET \cite{schick-schutze-2021-exploiting,schick-schutze-2021-just} utilizes the manually-designed prompts to reformulate many tasks as cloze questions (e.g., by appending phrases such as ``Similar sense of two sentences?") and performs gradient-based fine-tuning with smaller PLMs. To simplify PET/iPET, ADAPET \cite{tam-etal-2021-improving} decouples the losses for the label tokens and a label-conditioned masked language modeling objective over the full original input. Considering that manually designing the discrete prompts is time-consuming and labor-intensive, several efforts focus on searching proper discrete prompts automatically \cite{shin-etal-2020-autoprompt,gao2020making,zhong-etal-2021-factual}.\par
\begin{figure}[t]
    \centering
    \includegraphics[width=8.5cm]{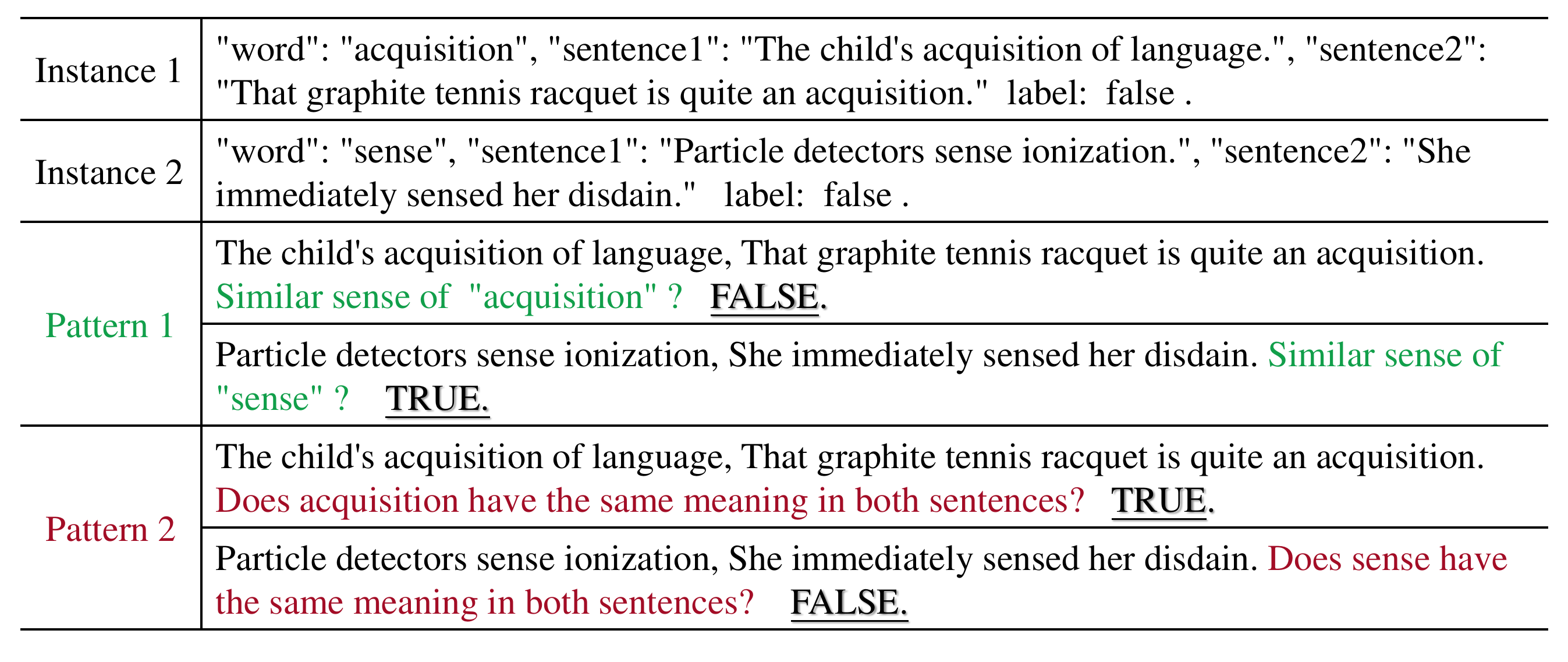}
    \caption{The example is chosen from WiC dataset in SuperGLUE. The color words indicate the manually-designed patterns which are used to formalize the instance into close-style questions.}
    \label{fig1:prompt-tuning}
\end{figure}
Although discrete prompts can reflect rationality from the perspective of humans, it may not be necessarily suitable for PLMs. To tackle this problem, a lot of studies begin to focus on continuous prompts. Continuous prompts can be thought of as special tokens. \cite{lester2021power} proposes prompt tuning and concatenates the continuous prompts with the embedding layer of PLMs. When using small PLMs, the performance of prompt tuning has a clear gap with fine-tuning. \cite{li-liang-2021-prefix} proposes prefix tuning and shows comparable results with fine-tuning on generation tasks. Prefix tuning concatenates continuous prompts with each layer in the decoder and only optimizes $0.1\% $ of the model parameters. However, the current usage of discrete and continuous prompts assumes that all samples in one task share the same prompt, and does not consider 
the diversity of the instances in which some are easy and others are difficult. Therefore, it is desirable to learn a special prompt for each instance.\par
In this paper, we propose an \textbf{I}nstance-aware \textbf{P}rompt \textbf{L}earning method (abbreviated as IPL) which learns a unique prompt for each instance. As shown in Figure \ref{fig1:prompt-tuning}, we use two different manually-designed patterns to formalize the instances into cloze-style questions and feed them into the pre-trained language model (PLM). As we can see, for each instance, using different prompts can get different answers. Pattern 1 is suitable for instance 1, while pattern 2 fits instance 2, which means that every instance needs a specific prompt for itself. However, it is difficult to dynamically find an appropriately discrete prompt for each instance. Therefore, we consider to utilize a look-up module and obtain a dynamic continuous prompt for each instance. Specifically, we take each learnable prompt token as a query and calculate its contribution to each instance through the look-up module. After doing this, each learnable prompt token has a different contribution to the instance and the contribution weighted continuous prompts are then utilized to guide the PLMs to perform the downstream task more instance-aware.\par 
We evaluate our approach on natural language understanding (NLU) and generation (NLG) tasks. For NLU tasks, we conduct experiments on SuperGLUE \cite{Wang2019SuperGLUEAS} with both GPT2 \cite{radford2019language} and RoBERTa \cite{Delobelle2020RobBERTAD}. For NLG tasks, we conduct experiments on table-to-text generation and summarization using GPT2. Experimental results on various tasks demonstrate that our method obtains considerable improvements compared to strong baselines. Especially, our method achieves the new state-of-the-art on the SuperGLUE few-shot learning benchmark. In summary, our key contributions can be listed as follows: \par
\begin{itemize}
\item {We propose an instance-aware prompt learning method that can learn a unique prompt for each instance.} \par
\item {Extensive experiments on both language understanding and generation tasks under both unidirectional and bidirectional PLMs verify the effectiveness of our method.}\par
\item {Detailed analyses verify that IPL can indeed dynamically learn appropriate continuous prompts for each instance.} \par
\end{itemize}

\begin{table*}[t]
\centering
\tiny
\resizebox{\textwidth}{!}{%
\renewcommand{\arraystretch}{1.0}
\tiny
\begin{tabular}{@{}lcccccccc|c@{}}
\toprule
{Method} &
  {\begin{tabular}[c]{@{}c@{}}{BoolQ}\\ Acc.\end{tabular}} &
  {\begin{tabular}[c]{@{}c@{}}CB\\ Acc./F1\end{tabular}} &
  {\begin{tabular}[c]{@{}c@{}}MultiRC\\ EM/F1a\end{tabular}} &
  {\begin{tabular}[c]{@{}c@{}}RTE\\ Acc.\end{tabular}} &
  {\begin{tabular}[c]{@{}c@{}}WiC\\ Acc.\end{tabular}} &
  {\begin{tabular}[c]{@{}c@{}}COPA\\ Acc.\end{tabular}} &
  {\begin{tabular}[c]{@{}c@{}}WSC\\ Acc.\end{tabular}} &
  {\begin{tabular}[c]{@{}c@{}}ReCoRD\\ Acc./F1\end{tabular}} &
  \multicolumn{1}{l}{Avg} \\ \midrule
GPT-3{\tiny $^{\dagger}$}  & 77.5          & 82.1/57.2          & 32.5/74.8          & 72.9          & 55.3           & 92.0          & 75.0             & \textbf{89.0}/90.1 & 73.2          \\
PET{\tiny $^{\dagger}$}    & 79.4          & 85.1/59.4          & 37.9/77.3          & 69.8          & 52.4           & \textbf{95.0} & 80.1           & 86.0/86.5          & 74.1          \\
iPET{\tiny $^{\dagger}$}   & \textbf{80.6} & \textbf{92.9}/92.4          & 33.0/74.0          & 74.0            & 52.2           & \textbf{95.0} & 80.1           & 86.0/86.5          & 76.8          \\
ADAPET{\tiny $^{\ddagger}$} & 80.3          & 89.3/86.8          & \textbf{39.2/80.1} & \textbf{76.5} & 54.4           & 89.0          & 81.7           & 85.4/92.1          & 77.3          \\
\hline
IPL   & 79.2          & \textbf{92.9/94.8} & 38.5/76.8        & 76.2          & \textbf{64.6} & 91.0          & \textbf{84.8} & 83.6/\textbf{91.1}        & \textbf{79.3} \\ \hline
\end{tabular}%
}
\caption{Few-shot learning (32 examples) on SuperGLUE validation set with ALBERT-xxlarge-v2. $\dagger$ indicates the results reported in \protect\cite{schick-schutze-2021-just}, and $\ddagger$ indicates the results reported in \protect\cite{tam-etal-2021-improving}. }
\label{table1}
\end{table*}

\begin{table*}[]
\centering
\resizebox{\textwidth}{!}{%
\renewcommand{\arraystretch}{1.0}
\tiny
\begin{tabular}{@{}lcccccccc|c@{}}
\toprule
{Method} &
  {\begin{tabular}[c]{@{}c@{}}BoolQ\\ Acc.\end{tabular}} &
  {\begin{tabular}[c]{@{}c@{}}CB\\ Acc./F1\end{tabular}} &
  {\begin{tabular}[c]{@{}c@{}}MultiRC\\ EM/F1a\end{tabular}} &
  {\begin{tabular}[c]{@{}c@{}}RTE\\ Acc.\end{tabular}} &
  {\begin{tabular}[c]{@{}c@{}}WiC\\ Acc.\end{tabular}} &
  {\begin{tabular}[c]{@{}c@{}}COPA\\ Acc.\end{tabular}} &
  {\begin{tabular}[c]{@{}c@{}}WSC\\ Acc.\end{tabular}} &
  {\begin{tabular}[c]{@{}c@{}}ReCoRD\\ Acc./F1\end{tabular}} &
  \multicolumn{1}{l}{Avg} \\ \midrule
GPT-3$^{\dagger}$ &
  76.4 &
  75.6/52.0 &
  30.5/75.4 &
  69.0 &
  49.4 &
  \textbf{92.0} &
  80.1 &
  \textbf{90.2/90.1} &
  71.8 \\
PET$^{\dagger}$ &
  79.1 &
  87.2 / 60.2 &
  \textbf{36.4 / 76.6} &
  67.2 &
  50.7 &
  90.8 &
  \textbf{88.4} &
  85.4 / 85.9 &
  74.0 \\
iPET$^{\dagger}$ &
  \textbf{81.2} &
  88.8/79.9 &
  31.7/74.1 &
  70.8 &
  49.3 &
  90.8 &
  \textbf{88.4} &
  85.4 / 85.9 &
  75.4 \\
ADAPET$^{\ddagger}$ &
  80.0 &
  \textbf{92.0}/82.3 &
  35.7 / 76.2 &
  \textbf{75.0} &
  53.5 &
  85.4 &
  85.6 &
  85.5 / 86.1 &
  76.0 \\
\hline
IPL &
  78.4 &
  \textbf{92.0/85.9} &
  {\color[HTML]{212121} 35.1/75.9} &
  74.9 &
  \cellcolor[HTML]{FFFFFF}\textbf{60.9} &
  85.6 &
  84.9 &
  83.5/84.3 &
  \textbf{76.6} \\ \hline
\end{tabular}
}
\caption{Few-shot learning (32 examples) on SuperGLUE test set with ALBERT-xxlarge-v2. $\dagger$ indicates the results reported in \protect\cite{schick-schutze-2021-just}, and $\ddagger$ indicates the results reported in \protect\cite{tam-etal-2021-improving}}
\label{table2}
\end{table*}
\section{Approach}
In this section, we present the details of our model IPL.
\begin{figure}[t]
    \centering
    \includegraphics[width=8cm]{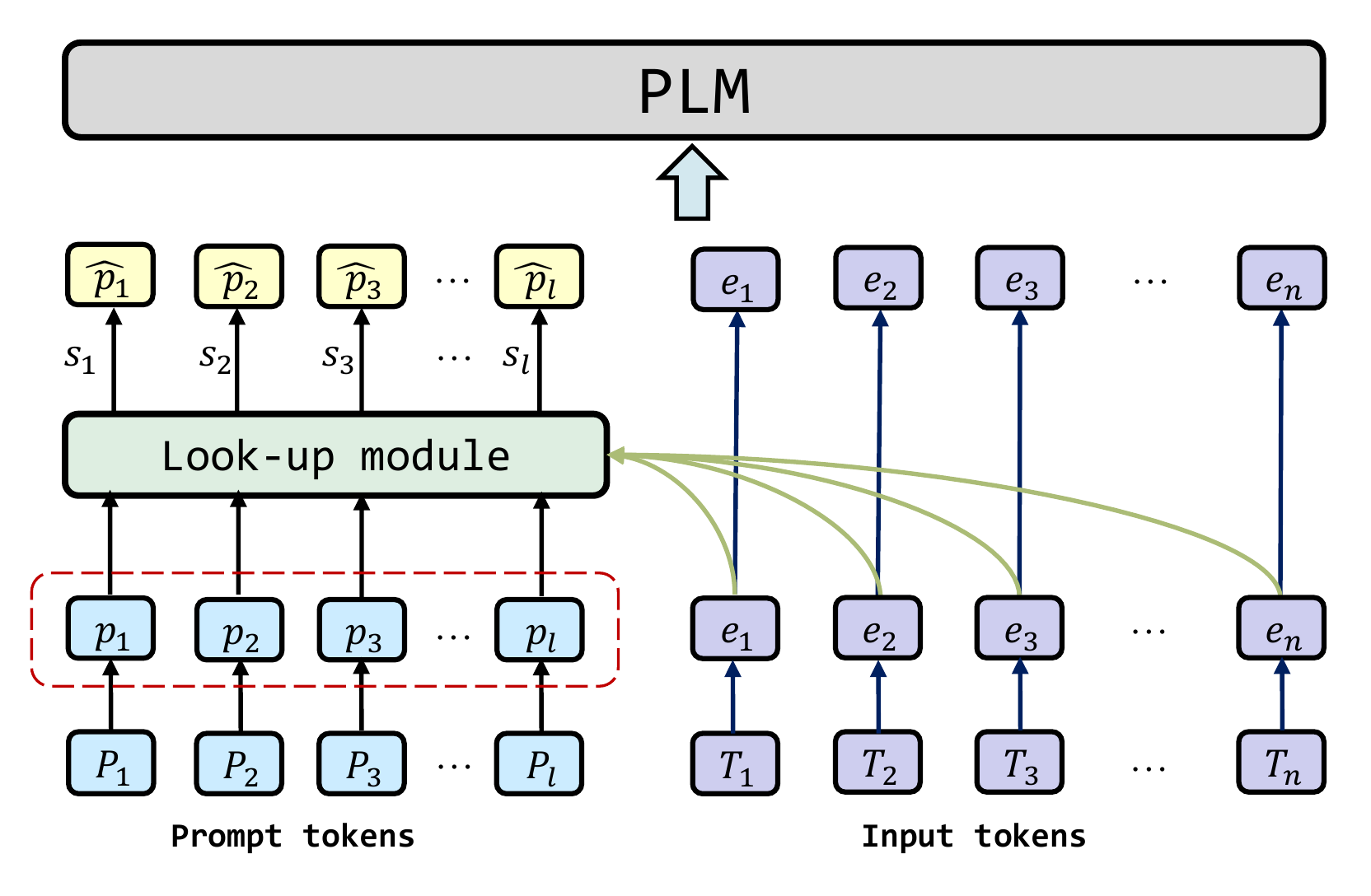}
    \caption{Illustration of our approach. $[P_1,P_2,P_3,\cdot\cdot\cdot,P_l]$ represents the prompt tokens. $[T_1,T_2,T_3,\cdot\cdot\cdot,T_n]$ represents the input tokens. $\{\bm{e_1},\bm{e_2},\bm{\cdot\cdot\cdot},\bm{e_n}\}$ represents the embeddings of input tokens. $\{\bm{p_1},\bm{p_2},\bm{\cdot\cdot\cdot},\bm{p_n}\}$ is the learnable prompt, and $[\bm{\hat {p_1}},\bm{\hat {p_2}},\bm{\cdot\cdot\cdot},\bm{\hat {p_l}}]$ refers to the weighted representation for the input instance.}
    \label{fig:Instance-aware prompt tuning}
\end{figure}
Previous studies demonstrate that prompt learning is promising for downstream tasks. However, using fixed prompts (e.g., discrete prompts like “convert the table into a sentence” or continuous prompts after optimization) for diverse instances in one task ignores the peculiarity of different instances. To address this problem, our instance-aware prompt learning method IPL can learn a special prompt for each specific instance. Next, we introduce prompt learning first and then present our IPL model.
\subsection{Prompt Learning}
For the standard paradigm of pre-training and fine-tuning, there is a gap (e.g., inconsistent objective function) between the pre-training stage and the fine-tuning stage. Fortunately, prompt learning bridges this gap by formalizing the downstream tasks into the form of a conditional language model or masked language model. Discrete prompt is an important method in prompt learning, for instance, given a masked language model $\mathcal{M}$, we first use the prompt to formulate a question and answer instance $x$ (e.g., [passage]. Can you have too much oxygen in your body ? ) as:
\begin{equation}\nonumber
\begin{aligned}
\hat{x} = x\ \text{the answer is [MASK]}.
\end{aligned}
\end{equation}
Then $\hat{x}$ is fed into $\mathcal{M}$, and let $\mathcal{M}$ determine whether "Yes" or "No" is more appropriate to replace [MASK] \cite{gao2020making}.\par
Continuous prompt is an alternative approach in prompt learning. Suppose we get the embedding sequence $\{\bm{e_1},\bm{e_2},\bm{\cdot\cdot\cdot},\bm{e_n}\}$ of the instance $x$, and we concatenate the continuous prompt (e.g., as shown in the red dotted box in Figure \ref{fig:Instance-aware prompt tuning}) $\{\bm{p_1},\bm{p_2},\bm{\cdot\cdot\cdot},\bm{p_l}\}$ with the embedding sequence, which can be formalized as follows:
\begin{equation}\nonumber
\begin{aligned}
\quad\quad\ \ \bm{\hat{x}} = \{\bm{p_1},\bm{p_2},\bm{\cdot\cdot\cdot},\bm{p_l};\bm{e_1},\bm{e_2},\bm{\cdot\cdot\cdot},\bm{e_n}\}
\end{aligned}
\end{equation}
Then, $\bm{\hat{x}}$ is fed into $\mathcal{M}$ to generate the target labels.\par
In this paper, we combine the advantages of continuous prompts with discrete prompts and propose an instance-aware prompt learning method which learns a unique prompt for each instance. Next, we detail our proposed IPL mdoel.
\subsection{Instance-aware Prompt Learning}\label{instance-aware}
We denote $\mathcal{T}$ as an instance in a form of a sequence consisting of n tokens $\mathcal{T}=\{T_1,T_2,\cdot\cdot\cdot,T_n\}$. We follow  \cite{li-liang-2021-prefix,lester2021power}, and use learnable prompts of special tokens $\mathcal{P} = \{P_1,P_2,\cdot\cdot\cdot,P_l\}$ and update the embeddings of these prompt tokens during prompt learning. As shown in Figure \ref{fig:Instance-aware prompt tuning}, we first use the pre-trained language model to obtain the embedding sequence $\{\bm{e_1},\bm{e_2},\bm{\cdot\cdot\cdot},\bm{e_n}\}$ of $\mathcal{T}$, and get a matrix $\bm{X}\in\mathbb{R}^{n\times d_e}$, where $n$ is the length of instance $\mathcal{T}$, $d_e$ is the dimension of the embedding space. Then we create a learnable matrix $\bm{P}=\{\bm{p_1},\bm{p_2},\bm{\cdot\cdot\cdot},\bm{p_l}\}$ of $\mathcal{P}$, where $l$ is the length of the prompt, and  $\bm{P}\in\mathbb{R}^{l\times d_e}$. After we get the prompt matrix $\bm{P}$ and embedding matrix $\bm{X}$, we use the projection matrices $\bm{W^M}\in\mathbb{R}^{d_e\times d_h}$ and $\bm{W^N}\in\mathbb{R}^{d_e\times d_h}$ to map $\bm{P}$ and $\bm{X}$ into matrix $\bm{M}\in\mathbb{R}^{l\times d_h}$ and $\bm{N}\in\mathbb{R}^{n\times d_h}$ respectively, where $d_h$ is the dimension of projection space.
\begin{equation}
\begin{split}
    \bm{M} &= \bm{P W^M} \\
    \bm{N} &= \bm{X W^N}
\end{split}
\end{equation}
\par
We suppose that each learnable prompt token has a different contribution to different instances and we learn the contribution scores by calculating the relevance score between matrix $\bm{M}=\{\bm{p_1^\prime},\bm{p_2^\prime},\bm{\cdot\cdot\cdot},\bm{p_l^\prime}\}$ and $\bm{N}=\{\bm{e_1^\prime},\bm{e_2^\prime},\bm{\cdot\cdot\cdot},\bm{e_n^\prime}\}$. After we get the relevance score, we pass the score to the look-up module. In the look-up module, we adopt a method of mean operation and apply a sigmoid function $\sigma$ to obtain how much does each learnable prompt token contributes to the instance $\mathcal{T}$. The detailed calculation is as follows:
\begin{gather}
    s_j = \sigma(\frac{1}{n}\sum_{i=1}^{n}\bm{p_j^\prime}\cdot\bm{{e_n^\prime}^T}) \\
    \bm{\hat p_j} = s_j\cdot \bm{p_j}
\end{gather}
where $\bm{{e_n^\prime} ^T}$ is the transpose of $\bm{e_n}$. $\bm{ p_j^{\prime}}$ is the contribution of the \emph{j}-th prompt token to the instance $\mathcal{T}$. $s_j$ is the contribution score of the \emph{j}-th prompt token after applying a sigmoid funtion $\sigma$, and $\bm{\hat {p_j}}$ is the \emph{j}-th weighted representation for the instance. After doing such a calculation for all prompt tokens, we get the weighted prompt as $\bm{\hat{P}}=\{\bm{\hat {p_1}},\bm{\hat {p_2}},\bm{\cdot\cdot\cdot},\bm{\hat {p_l}}\}$. Then we concatenate weighted continuous prompt with the embedded instance as a new matrix $[\bm{\hat{P}};\bm{X}]\in\mathbb{R}^{(l+n)\times d_e}$, and feed it into the pre-trained language model.\par
During training, we optimize the parameters of prompt module and PLMs. We do not freeze the model parameters as \cite{lester2021power} does, because their results demonstrates that the gap between prompt tuning and fine-tuning disappear only when the model size increases to 10 billion parameters.
\begin{table*}[]
\centering
\resizebox{\textwidth}{!}{%
\tiny
\renewcommand{\arraystretch}{1.2}
\begin{tabular}{@{}ccccccl|lccccc@{}}
\toprule
\textbf{Method} &
  \begin{tabular}[c]{@{}c@{}}BoolQ\\ Acc.\end{tabular} &
  \begin{tabular}[c]{@{}c@{}}CB\\ Acc./F1\end{tabular} &
  \begin{tabular}[c]{@{}c@{}}MultiRC\\ F1a/EM\end{tabular} &
  \begin{tabular}[c]{@{}c@{}}RTE\\ Acc.\end{tabular} &
  \begin{tabular}[c]{@{}c@{}}WiC\\ Acc.\end{tabular} &
   &
   &
  \begin{tabular}[c]{@{}c@{}}BoolQ\\ Acc.\end{tabular} &
  \begin{tabular}[c]{@{}c@{}}CB\\ Acc./F1\end{tabular} &
  \begin{tabular}[c]{@{}c@{}}MultiRC\\ F1a/EM\end{tabular} &
  \begin{tabular}[c]{@{}c@{}}RTE\\ Acc.\end{tabular} &
  \begin{tabular}[c]{@{}c@{}}WiC\\ Acc.\end{tabular} \\ \hline \hline
\multicolumn{1}{c}{} &
  \multicolumn{1}{c}{} &
  \multicolumn{1}{c}{} &
  \texttt{GPT2-base} &
  \multicolumn{1}{c}{} &
  \multicolumn{1}{c}{} &
   &
   &
  \multicolumn{1}{c}{} &
  \multicolumn{1}{c}{} &
  \texttt{GPT2-large} &
  \multicolumn{1}{c}{} &
  \multicolumn{1}{c}{} \\
\hline
\hline
PET &
  74.55 &
  94.05/95.58 &
  {70.36/\textbf{22.87}} &
  67.14 &
  65.67 &
   &
   &
  80.43 &
  92.86/94.75 &
  75.81/32.42 &
  78.70 &
  \textbf{70.22} \\
PT &
  74.16 &
  92.86/94.70 &
  69.84/21.16 &
  67.03 &
  64.16 &
   &
   &
  79.66 &
  96.43/97.37 &
  75.81/\textbf{34.10} &
  75.45 &
  69.12 \\
\hline
IPL &
  \textbf{74.89} &
  \textbf{94.64/96.03} &
  \textbf{70.54}/22.35 &
  \textbf{69.68} &
  \textbf{66.72} &
   &
   &
  \textbf{80.80} &
  \textbf{98.21/98.67} &
  \textbf{76.00}/33.26 &
  \textbf{80.14} &
  69.59 \\ \hline
\end{tabular}%
}
\caption{Fully-supervised learning on SuperGLUE validation set with unidirectional pre-trained language models. PET means PET fine-tuning with a single pattern, and PT refers to prompt-tuning. For a fair comparison, we use the same pattern for all models.}
\label{table3}
\end{table*}
\begin{table*}[]
\centering
\resizebox{\textwidth}{!}{%
\tiny
\renewcommand{\arraystretch}{1.2}
\begin{tabular}{@{}ccccccl|lccccc@{}}
\toprule
\textbf{Method} &
  \begin{tabular}[c]{@{}c@{}}BoolQ\\ Acc.\end{tabular} &
  \begin{tabular}[c]{@{}c@{}}CB\\ Acc./F1\end{tabular} &
  \begin{tabular}[c]{@{}c@{}}MultiRC\\ F1a/EM\end{tabular} &
  \begin{tabular}[c]{@{}c@{}}RTE\\ Acc.\end{tabular} &
  \begin{tabular}[c]{@{}c@{}}WiC\\ Acc.\end{tabular} &
   &
   &
  \begin{tabular}[c]{@{}c@{}}BoolQ\\ Acc.\end{tabular} &
  \begin{tabular}[c]{@{}c@{}}CB\\ Acc./F1\end{tabular} &
  \begin{tabular}[c]{@{}c@{}}MultiRC\\ F1a/EM\end{tabular} &
  \begin{tabular}[c]{@{}c@{}}RTE\\ Acc.\end{tabular} &
  \begin{tabular}[c]{@{}c@{}}WiC\\ Acc.\end{tabular} \\ \hline\hline
\multicolumn{1}{c}{} &
  \multicolumn{1}{c}{} &
  \multicolumn{1}{c}{} &
  \texttt{RoBERTa-base} &
  \multicolumn{1}{c}{} &
  \multicolumn{1}{c}{} &
   &
   &
  \multicolumn{1}{c}{} &
  \multicolumn{1}{c}{} &
  \texttt{RoBERTa-large} &
  \multicolumn{1}{c}{} &
  \multicolumn{1}{c}{} \\
\hline
\hline
PET &
  80.03 &
  \textbf{96.43/95.56} &
  76.05/\textbf{35.12} &
  82.67 &
  69.28 &
   &
   &
  85.47 &
  98.81/99.12 &
  83.36/\textbf{51.07} &
  87.12 &
  70.85 \\
PT &
  80.32 &
  96.43/94.76 &
  76.05/33.44 &
  80.14 &
  68.86 &
   &
   &
  85.44 &
  98.81/99.12 &
  83.23/50.75 &
  87.01 &
  72.14 \\
\hline
IPL &
  \textbf{80.63} &
  \textbf{96.43/95.56} &
  \textbf{76.19}/34.59 &
  \textbf{82.91} &
  \textbf{70.85} &
   &
   &
  \textbf{85.66} &
  \textbf{99.40/99.56} &
  {\textbf{83.40}/50.89}&
  \textbf{87.48} &
  \textbf{73.51} \\ \hline
\end{tabular}%
}
\caption{Fully-supervised learning on SuperGLUE validation set with bidirectional pre-trained language models. PET means PET fine-tuning with a single pattern, and PT refers to prompt tuning. For a fair comparison, we use the same pattern for all the models.}
\label{table4}
\end{table*}
\section{Experiments}
To evaluate our method IPL, we conduct experiments on both NLU tasks and NLG tasks. For NLU tasks, we evaluate IPL on SuperGLUE$\footnote{\url{https://supergluebenchmark.com/}}$ \cite{Wang2019SuperGLUEAS}. And we evaluate IPL for few-shot learning by using 32 labeled examples per task from FewGLUE\footnote{\url{https://github.com/timoschick/fewglue}} \cite{schick-schutze-2021-just}). For NLG tasks, we select 3 standard table-to-text generation tasks: E2E \cite{Novikova2017TheED}, WebNLG \cite{Gardent2017TheWC}, DART \cite{Radev2021DARTOS} and a dialogue summarization task: SamSum \cite{Gliwa2019SAMSumCA}.
 \subsection{Experiments on NLU Tasks}
 Our code is implemented based on PET$\footnote{\url{https://github.com/timoschick/pet}}$ using HuggingFace \cite{Wolf2020TransformersSN}.
The experiment results include the few-shot learning results and fully-supervised learning results.
\subsubsection{Few-shot Learning Results}
 For a fair comparison, we choose ALBERT-xxlarge-v2 \cite{Lan2020ALBERTAL} for experiments and use the same data split as \cite{schick-schutze-2021-just}, which consists of 32 labeled examples for each task. We use a default setting training for 20 epochs, using a learning rate of 1e-5, a batch size of 8, and a prompt length of 16.\par
Our main results on the validation and test sets on SuperGLUE are shown in Table \ref{table1} and Table \ref{table2}. We compare against GPT-3, PET/iPET and ADAPET. Initially, ADAPET does not use the unlabeled data and achieves the state-of-the-art on SuperGLUE few-shot learning tasks compared to PET/iPET which uses the unlabeled data. And for IPL, we train IPL with a single pattern and do not use the unlabeled data.\par
As can be seen from Table \ref{table1}, on average, IPL outperforms GPT-3 by 6 points; outperforms PET's iterative variant, iPET, by 2.5 points, and even outperforms the previous state-of-the-art model ADAPET by 2 points on the dev set. Specifically, compared with iPET and GPT-3, IPL achieves improvements on 5 out of the 8 tasks and 6 out of the 8 tasks respectively, demonstrating the effectiveness of our method in few-shot NLU tasks. We will conduct detailed analysis in \ref{Visualization of Instance Aware Prompt}.\par
We also report the test set on SuperGLUE in Table \ref{table2}, IPL outperforms all other models including ADAPET, which is the previous state-of-the-art model, and obtains the new state-of-the-art for few-shot learning on SuperGLUE.\par
\subsubsection{Fully-supervised Learning Results}
To verify IPL on various PLMs, we perform experiments on 5 out of the 8 tasks of SuperGLUE benchmark including BoolQ, MultiRC, RTE, CB, and WiC, where we choose both unidirectional PLMs GPT-2 and bidirectional PLMs RoBERTa. We report the performance of fine-tuning with PET \cite{schick-schutze-2021-exploiting}, prompt tuning \cite{lester2021power}, and our method IPL. We use a default setting training for 20 epochs, using a learning rate of 2e-5, a batch size of 32, and a prompt length of 16.\par
Table \ref{table3} and Table \ref{table4} show our main results on GPT-2 and RoBERTa. On unidirectional PLMs like GPT2-base and GPT2-large, IPL outperforms PET fine-tuning and prompt tuning on all 5 tasks with GPT2-base and 4 out of 5 tasks on GPT2-large. On bidirectional PLMs like RoBERTa-base and RoBERTa-large, IPL outperforms all other RoBERTa-based models on all 5 tasks. Based on the experiment results, we demonstrate that IPL can achieve great results on both GPT-2 and RoBERTa models.
\subsection{Experiment on NLG Tasks}
For NLG tasks, we compare IPL on GPT2-base and GPT2-large with two baseline methods: the standard fine-tuning, and prompt tuning, where we do not freeze the model parameters as IPL does. The experiment results are illustrated in Table \ref{table5} and Table \ref{table6}. We choose three table-to-text tasks and a summarization task. For table-to-text tasks, on E2E, we use the official evaluation script, which reports BLUE \cite{Papineni2002BleuAM}, NIST \cite{Belz2006ComparingAA}, ROUGE-L \cite{Lin2004ROUGEAP}, and CIDEr \cite{Vedantam2015CIDErCI}. On WebNLG, we use the official evaluation script, which reports BLEU, METEOR \cite{Lavie2007METEORAA}, and TER \cite{snover-etal-2006-study}. On DART, we use the official evaluation script and report BLEU, METEOR, TER, and BERTScore \cite{Zhang*2020BERTScore:}. As for the summarization task: SamSum, we report ROUGE-1, ROUGE-2, and ROUGE-L. The hyperparameters we tune include the number of epochs, batch size, learning rate, and prefix length. For table-to-text tasks, we set batch size as 32, prefix length as 10, the number of epochs as 10 for both GPT2-base and GPT2-large, in addition to the learning rate as 5e-5 for GPT-base, 5e-6 for GPT-large. For the summarization task, we set the prefix length as 100 and other hyperparameters keep consistent with table-to-text tasks.
\begin{table*}[]
\centering
\resizebox{\textwidth}{!}{%
\small
\renewcommand{\arraystretch}{1.4}
\begin{tabular}{@{}ccccc|ccc|ccc|ccc|cccc@{}}
\toprule
 &
  \multicolumn{4}{c}{E2E} &
  \multicolumn{9}{c}{WebNLG} &
  \multicolumn{4}{c}{DART} \\ 
\multicolumn{1}{l}{} &
  \multicolumn{1}{l}{BLUE} &
  \multicolumn{1}{l}{NIST} &
  \multicolumn{1}{l}{R\_L} &
  \multicolumn{1}{l}{CIDEr} &
  \multicolumn{3}{c}{BLUE} &
  \multicolumn{3}{c}{MET} &
  \multicolumn{3}{c}{TER $\downarrow$} &
  \multicolumn{1}{l}{BLUE} &
  \multicolumn{1}{l}{MET} &
  \multicolumn{1}{l}{TER $\downarrow$} &
  \multicolumn{1}{l}{BERT} \\ \hline
 &
   &
   &
   &
   &
  Seen &
  Unseen &
  All &
  Seen &
  Unseen &
  All &
  Seen &
  Unseen &
  All &
   &
   &
   &
  \multicolumn{1}{l}{} \\
\midrule\midrule
\multicolumn{1}{l}{} &
  \multicolumn{1}{l}{} &
  \multicolumn{1}{l}{} &
  \multicolumn{1}{l}{} &
  \multicolumn{1}{l}{} &
  \multicolumn{1}{l}{} &
  \multicolumn{1}{l}{} &
  \multicolumn{1}{l}{} &
  \multicolumn{1}{l}{} &
  \multicolumn{1}{l}{\texttt{GPT2-base}} &
  \multicolumn{1}{l}{} &
  \multicolumn{1}{l}{} &
  \multicolumn{1}{l}{} &
  \multicolumn{1}{l}{} &
  \multicolumn{1}{l}{} &
  \multicolumn{1}{l}{} &
  \multicolumn{1}{l}{} &
  \multicolumn{1}{l}{} \\
\midrule\midrule
FT &
  69.55 &
  8.79 &
  71.52 &
  \textbf{2.49} &
  56.01 &
  26.46 &
  41.70 &
  39.33 &
  25.04 &
  32.46 &
  46.40 &
  81.80 &
  62.63 &
  42.08 &
  35.17 &
  52.58 &
  94.12 \\
PT &
  69.78 &
  8.81 &
  71.55 &
  \textbf{2.49} &
  60.55 &
  28.03 &
  45.51 &
  43.25 &
  28.82 &
  36.30 &
  38.03 &
  74.17 &
  54.60 &
  \textbf{45.27} &
  \textbf{37.62} &
  49.83 &
  94.76 \\
\hline
IPL &
  \textbf{69.82} &
  \textbf{8.82} &
  \textbf{71.65} &
  \textbf{2.49} &
  \textbf{60.93} &
  \textbf{29.94} &
  \textbf{46.46} &
  \textbf{43.27} &
  \textbf{29.15} &
  \textbf{36.50} &
  \textbf{37.76} &
  \textbf{72.23} &
  \textbf{53.56} &
  42.98 &
  35.62 &
  \textbf{48.50} &
  \textbf{95.43} \\
\midrule\midrule
\multicolumn{1}{l}{} &
  \multicolumn{1}{l}{} &
  \multicolumn{1}{l}{} &
  \multicolumn{1}{l}{} &
  \multicolumn{1}{l}{} &
  \multicolumn{1}{l}{} &
  \multicolumn{1}{l}{} &
  \multicolumn{1}{l}{} &
  \multicolumn{1}{l}{} &
  \multicolumn{1}{l}{\texttt{GPT2-large}} &
  \multicolumn{1}{l}{} &
  \multicolumn{1}{l}{} &
  \multicolumn{1}{l}{} &
  \multicolumn{1}{l}{} &
  \multicolumn{1}{l}{} &
  \multicolumn{1}{l}{} &
  \multicolumn{1}{l}{} &
  \multicolumn{1}{l}{} \\
\midrule\midrule
FT &
  68.32 &
  \textbf{8.76} &
  \textbf{71.25} &
  2.48 &
  62.11 &
  43.52 &
  53.61 &
  44.56 &
  37.39 &
  41.21 &
  37.06 &
  53.62 &
  44.65 &
  47.16 &
  38.24 &
  47.35 &
  94.43 \\
PT &
  68.32 &
  8.65 &
  71.04 &
  2.49 &
  \textbf{64.18} &
  46.04 &
  55.85 &
  \textbf{45.30} &
  38.62 &
  \textbf{42.17} &
  \textbf{34.81} &
  50.92 &
  \textbf{42.19} &
  \textbf{48.57} &
  39.04 &
  \textbf{46.12} &
  94.90 \\
\hline
IPL &
  \textbf{68.53} &
  8.68 &
  71.2 &
  \textbf{2.51} &
  64.06 &
  \textbf{46.12} &
  \textbf{55.90} &
  45.24 &
  \textbf{38.64} &
  42.12 &
  35.28 &
  \textbf{50.55} &
  42.28 &
  48.38 &
  \textbf{39.15} &
  46.17 &
  \textbf{95.47} \\ \hline
\end{tabular}%
}
\caption{The best score is in bold for both GPT2-base and GPT2-large. The FT refers to fine-tuning. PT refers to prompt tuning. For the metrics, the higher the better except for TER.}
\label{table5}
\end{table*}
\subsubsection{Results on NLG Tasks}
As shown in Table \ref{table5}, on GPT2-base, IPL performs better than fine-tuning and prompt tuning on E2E and WebNLG, while on DART, which is an open domain table-to-text dataset, IPL slightly underperforms prompt tuning. On GPT2-large, IPL outperforms fine-tuning and can be comparable or better than prompt tuning. Additionally, IPL obtains better performance on WebNLG unseen domains suggesting that IPL can generalize to other domains better. For the summarization task, the results in Table \ref{table6} show IPL performs better than fine-tuning and prompt tuning on both GPT2-base and GPT2-large models, suggesting it has the potential to scale to even larger models. Above all, the results demonstrate that IPL can achieve great results on NLG tasks with GPT-2 model. \par
\begin{figure}[t]
\centering
\subfigure[Prompt length (Few-shot)]{
\begin{minipage}[t]{0.5\linewidth}
\centering
\includegraphics[width=4cm]{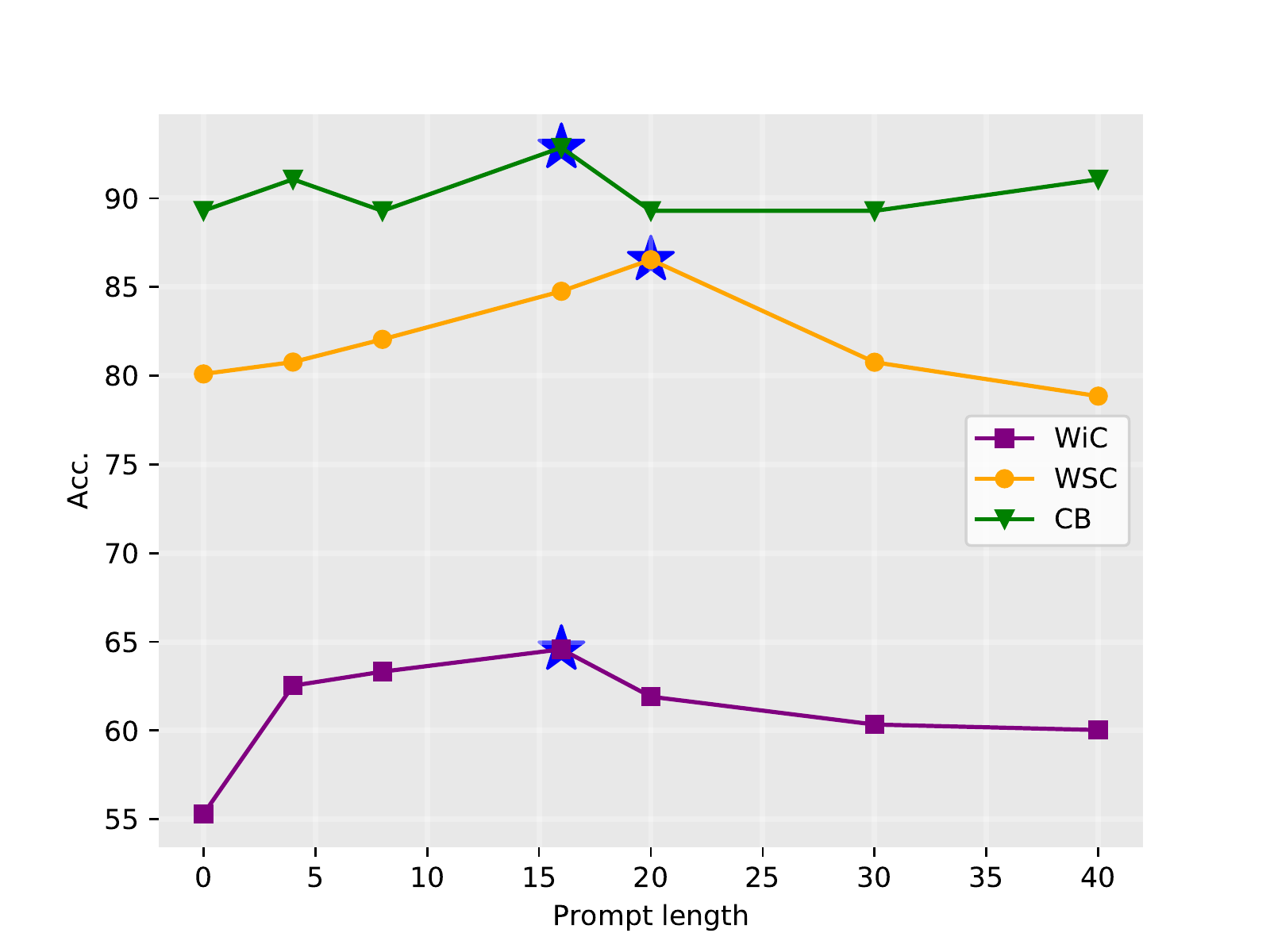}
\label{3a}
\end{minipage}%
}%
\subfigure[Prompt length (SamSum)]{
\begin{minipage}[t]{0.5\linewidth}
\centering
\includegraphics[width=4cm]{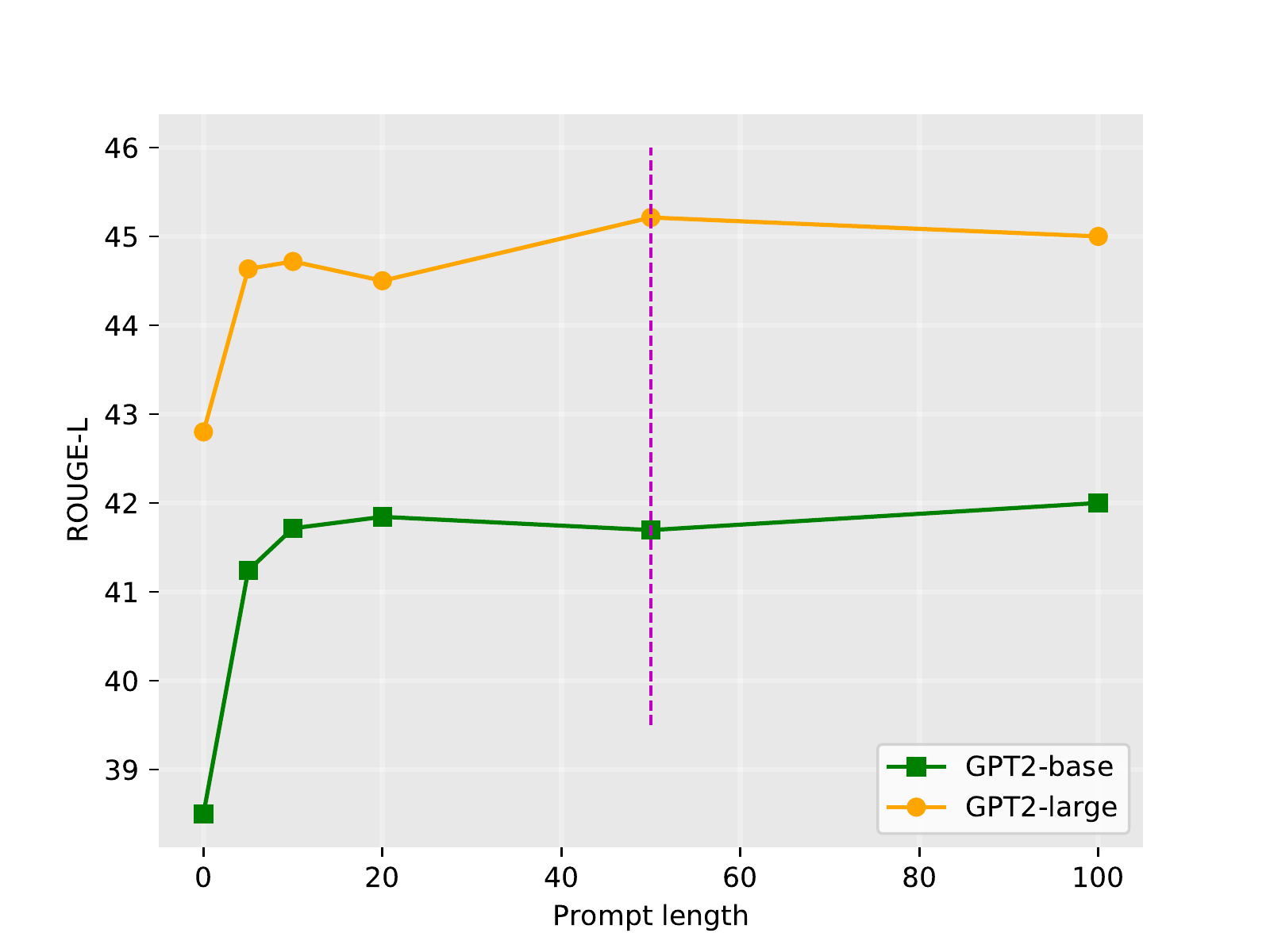}
\label{3b}
\end{minipage}%
}%
\caption{(a) Performance on validation set in the few-shot of SuperGLUE including WiC, WSC, and CB, while the prompt length varies in \{0,4,8,16,20,30,40\}. (b) Performance on SamSum while the prompt length varies in \{0,5,10,20,50,100\}.}
\end{figure}

\begin{figure}[t]
    \centering
    \includegraphics[width=8.8cm]{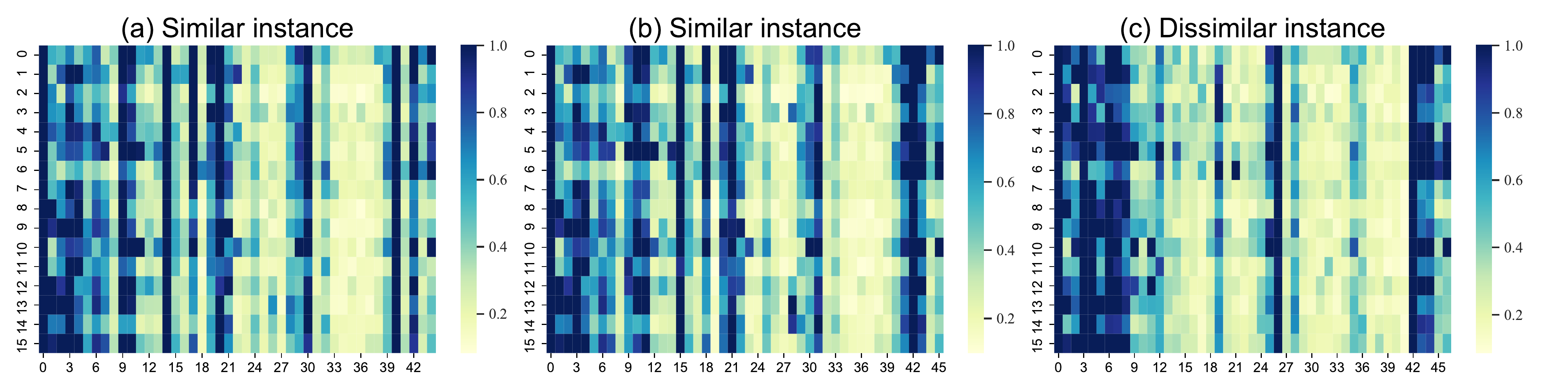}
    \caption{Attention visualization of different instances. (a) and (b) are similar instances, (a) and (c) or (b) and (c) are dissimilar instances. X-axis represents the input sequence without prompt, and Y-axis represents the prompt sequence.}
    \label{case_attention} 
    
\end{figure}

\section{Analysis}\label{Analysis}
We conduct detailed analyses on IPL. Section \ref{prompt length} studies the effect of the prompt length on the performance of few-shot learning on SuperGLUE. Section \ref{Visualization of Instance Aware Prompt} visualizes the attention matrix of similar instances and dissimilar instances to verify the effectiveness of our approach. Section \ref{case_study} shows the results of different methods on different instances.
\subsection{Prompt Length}\label{prompt length}
We visualize the relationship between the performance and different prompt lengths (other settings are fixed). For NLU tasks, we conduct experiments on three tasks of SuperGLUE including CB, WSC, WiC using ALBERT-xxlarge-v2. Figure \ref{3a} shows that performance increases as the prompt length increases up to a threshold (16 for CB and WiC, 20 for WSC), and then the performance slightly drops. Figure \ref{3b} shows the effect of prompt length on the performance of different model sizes on SamSum. We can see that the performance consistently increases until the prompt length is up to 50. Continuing to increase the prompt length cannot yield significant improvements. 

\subsection{Visualization of Instance-aware Prompt}\label{Visualization of Instance Aware Prompt}
We choose similar instances and dissimilar instances from WSC \cite{Levesque2011TheWS} for analysis. Figure \ref{case_attention} shows the analysis results for IPL on similar instances and dissimilar instances. We visualize the attention matrix of the instance and prompt. As shown in \ref{case_attention}(a) and \ref{case_attention}(b), the attention matrices between \ref{case_attention}(a) and \ref{case_attention}(b) are similar, which means IPL can produce similar prompts for similar instances. Comparing \ref{case_attention}(a) and \ref{case_attention}(c) or \ref{case_attention}(b) and \ref{case_attention}(c), we find that the attention matrices are not similar, suggesting that IPL can produce different prompts for dissimilar instances. Consequently, our approach learns a special prompt for each instance and can be aware of the important information of the instance.\par
\subsection{Case study}\label{case_study}
\begin{figure}[t]
    \centering
    \includegraphics[width=8.8cm]{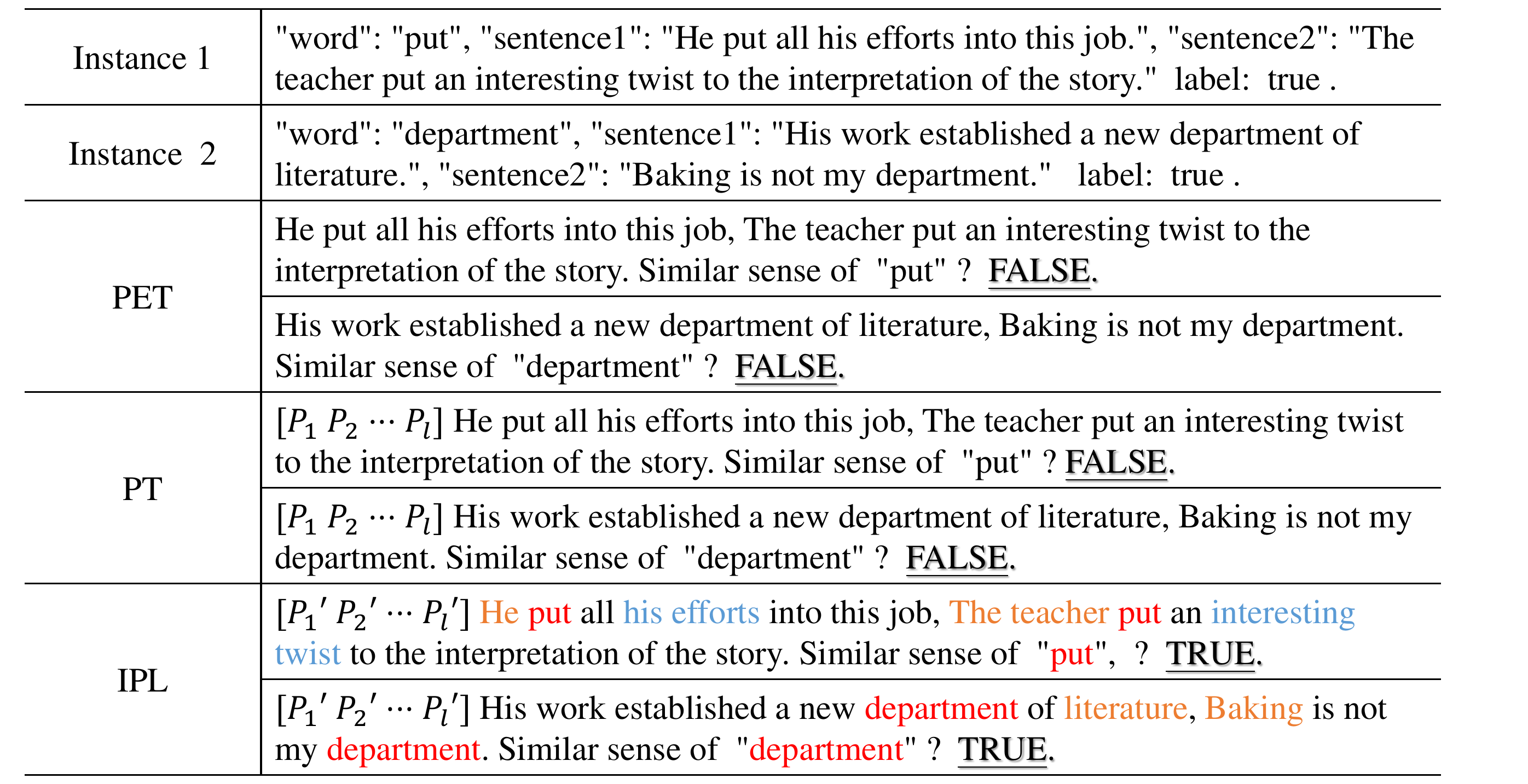}
    \caption{The instances are chosen from WiC dataset in SuperGLUE, which is shown on the top. We use the manually-designed pattern from PET. PT refers to prompt tuning. Our method is shown on the bottom, and the color words mean our approach can be aware of the important words in the instance.
}
    \label{figure5}
\end{figure}

As shown in Figure \ref{figure5}, for indistinguishable instances, PET utilizes a fixed discrete prompt and makes a wrong judgment on the meaning of the word `put' and `department'. Prompt tuning prepends the fixed continuous prompt with the two instances based on the pattern of PET and also gives wrong answers. In contrast, our method IPL learns a unique prompt for each instance and contains much information of the instance yielding the correct answer.
\begin{table}[]
\centering
\small
\begin{tabular}{@{}cccc|ccc@{}}
\toprule
Method & R-1  & R-2  & R-L  & R-1  & R-2  & R-L  \\ \midrule\midrule
\multicolumn{1}{l}{} & \multicolumn{3}{c}{\texttt{GPT2-base}}                 & \multicolumn{3}{c}{\texttt{GPT2-large}}                \\ \midrule\midrule
FT     & 42.6 & 18.9 & 38.5 & 47.2 & 22.2 & 42.8 \\
PT     & 46.5 & 21.4 & 41.8 & 49.3 & 24.5 & 44.8 \\ \midrule
IPL                 & \textbf{46.6} & \textbf{21.7} & \textbf{42.0} & \textbf{49.7} & \textbf{24.8} & \textbf{45.0} \\ \midrule
\end{tabular}
\caption{Results for summarization on SamSum using GPT2 models. The FT refers to fine-tuning. PT refers to prompt tuning, and the best score is in bold.}
\label{table6}
\end{table}
\section{Related Work}
GPT-3 \cite{Brown2020LanguageMA}, which uses the task description and several typical examples as prompt to guide the generation, indicates the language models are few-shot learners and leads to the waves of prompt learning. Recently, PET/iPET \cite{schick-schutze-2021-just} utilizes the manually-designed prompts to reformulate natural language understanding tasks as cloze-style questions with gradient-based fine-tuning. 
There are also a lot of studies that utilize the manually-designed prompt to mine the knowledge from the PLMs \cite{jiang-etal-2020-know,Trinh2018ASM}.
Since manual-designed prompt is time-consuming and the search space is huge, researches focus on automatic prompt search \cite{gao2020making,shin-etal-2020-autoprompt,zhong-etal-2021-factual}. 

However, the handcrafted prompt can only reflect rationality from the perspective of humans, which midwifery the exploration in continuous prompts. \cite{li-liang-2021-prefix} proposes prefix tuning and concatenates learnable prompt at each layer of transformer while only optimizing the prefix parameters. In contrast, prompt tuning \cite{lester2021power} concatenates learnable prompt only in the embedding layer and only optimizes the prompt parameters in the embedding layer. Although \cite{lester2021power} demonstrates the effectiveness of light-weight prompt-tuning, the gap with full parameter fine-tuning still exists especially when the PLM is small. 

There are also a lot of works that interleave the prompt throughout the input layer. \cite{hambardzumyan-etal-2021-warp} proposes WARP, initializing the prompt parameters either with word embeddings of [MASK] or similar to the vectors from the word embedding layer. Their work is based on a series of masked language models \cite{Delobelle2020RobBERTAD,Lan2020ALBERTAL} and uses a learnable output layer to project the mask to class logits, which restricts the model and only produces a single output. \cite{Liu2021GPTUT} proposes P-tuning, using the patterns based on human design and putting the continuous prompts interleave throughout the embedded input. When optimizing the model, P-tuning jointly updates both the prompt and model parameters to perform on SuperGLUE. Similarly, we borrow the idea of human designed patterns to convert different tasks into the form of conditional language model or masked language model, and also apply our method on GPT-2 and RoBERTa.

However, the above usage of the discrete and continuous prompts assumes that the prompt is fixed for a specific task and all samples in the task share the same prompt. Different from the above methods, our proposed IPL dynamically learns a special prompt for each instance and obtains considerable improvements compared to strong baselines. 

Very recently, a contemporaneous work also presents another instance dependent prompt generation approach \cite{IDPG}, which studies only masked language model on only NLU tasks. In contrast, our IPL model is simple and effective for both unidirectional and bidirectional PLMs on both NLU and NLG tasks.\par

\section{Conclusion and Future Work}
In this paper, we propose an instance-aware prompt learning method named IPL, which learns a unique prompt for each instance. We find that IPL has the potential to be applied to both unidirectional and bidirectional PLMs on both language understanding and generation tasks. In the few-shot learning SuperGLUE benchmark, IPL outperforms all other models and obtains the new state-of-the-art. Detailed analysis demonstrates that our IPL model can indeed dynamically learn appropriate prompts for various instances.\par
In the future, we would explore how to generate better instance-aware prompts, and apply the instance-aware method to parameter-efficient tuning for more natural language processing tasks. 
\bibliographystyle{named}

\bibliography{ijcai22}

\begin{thebibliography}{}

\bibitem[\protect\citeauthoryear{Anonymous}{2021}]{IDPG}
Anonymous.
\newblock Idpg: An instance-dependent prompt generation method.
\newblock In {\em Openreview for ACL Rolling Review - November Submission},
  2021.

\bibitem[\protect\citeauthoryear{Belz and Reiter}{2006}]{Belz2006ComparingAA}
Anja Belz and Ehud Reiter.
\newblock Comparing automatic and human evaluation of nlg systems.
\newblock In {\em EACL}, 2006.

\bibitem[\protect\citeauthoryear{Brown \bgroup \em et al.\egroup
  }{2020}]{Brown2020LanguageMA}
Tom Brown, Benjamin Mann, Nick Ryder, Melanie Subbiah, Jared~D Kaplan, Prafulla
  Dhariwal, Arvind Neelakantan, Pranav Shyam, Girish Sastry, Amanda Askell,
  Sandhini Agarwal, Ariel Herbert-Voss, Gretchen Krueger, Tom Henighan, Rewon
  Child, Aditya Ramesh, Daniel Ziegler, Jeffrey Wu, Clemens Winter, Chris
  Hesse, Mark Chen, Eric Sigler, Mateusz Litwin, Scott Gray, Benjamin Chess,
  Jack Clark, Christopher Berner, Sam McCandlish, Alec Radford, Ilya Sutskever,
  and Dario Amodei.
\newblock Language models are few-shot learners.
\newblock In {\em NeurIPS}, 2020.

\bibitem[\protect\citeauthoryear{Delobelle \bgroup \em et al.\egroup
  }{2020}]{Delobelle2020RobBERTAD}
Pieter Delobelle, Thomas Winters, and Bettina Berendt.
\newblock {R}ob{BERT}: a {D}utch {R}o{BERT}a-based {L}anguage {M}odel.
\newblock In {\em Findings of EMNLP}, 2020.

\bibitem[\protect\citeauthoryear{Gao \bgroup \em et al.\egroup
  }{2021}]{gao2020making}
Tianyu Gao, Adam Fisch, and Danqi Chen.
\newblock Making pre-trained language models better few-shot learners.
\newblock In {\em ACL}, 2021.

\bibitem[\protect\citeauthoryear{Gardent \bgroup \em et al.\egroup
  }{2017}]{Gardent2017TheWC}
Claire Gardent, Anastasia Shimorina, Shashi Narayan, and Laura
  Perez-Beltrachini.
\newblock The {W}eb{NLG} challenge: Generating text from {RDF} data.
\newblock In {\em ICNLG}, 2017.

\bibitem[\protect\citeauthoryear{Gliwa \bgroup \em et al.\egroup
  }{2019}]{Gliwa2019SAMSumCA}
Bogdan Gliwa, Iwona Mochol, Maciej Biesek, and Aleksander Wawer.
\newblock {SAMS}um corpus: A human-annotated dialogue dataset for abstractive
  summarization.
\newblock In {\em Proceedings of the 2nd Workshop on New Frontiers in
  Summarization}, 2019.

\bibitem[\protect\citeauthoryear{Hambardzumyan \bgroup \em et al.\egroup
  }{2021}]{hambardzumyan-etal-2021-warp}
Karen Hambardzumyan, Hrant Khachatrian, and Jonathan May.
\newblock {WARP}: {W}ord-level {A}dversarial {R}e{P}rogramming.
\newblock In {\em ACL}, 2021.

\bibitem[\protect\citeauthoryear{Jiang \bgroup \em et al.\egroup
  }{2020}]{jiang-etal-2020-know}
Zhengbao Jiang, Frank~F. Xu, Jun Araki, and Graham Neubig.
\newblock How can we know what language models know?
\newblock {\em TACL}, 2020.

\bibitem[\protect\citeauthoryear{Lan \bgroup \em et al.\egroup
  }{2020}]{Lan2020ALBERTAL}
Zhenzhong Lan, Mingda Chen, Sebastian Goodman, Kevin Gimpel, Piyush Sharma, and
  Radu Soricut.
\newblock Albert: A lite bert for self-supervised learning of language
  representations.
\newblock In {\em ICLR}, 2020.

\bibitem[\protect\citeauthoryear{Lavie and Agarwal}{2007}]{Lavie2007METEORAA}
Alon Lavie and Abhaya Agarwal.
\newblock {METEOR}: An automatic metric for {MT} evaluation with high levels of
  correlation with human judgments.
\newblock In {\em Proceedings of the Second Workshop on Statistical Machine
  Translation}, 2007.

\bibitem[\protect\citeauthoryear{Lester \bgroup \em et al.\egroup
  }{2021}]{lester2021power}
Brian Lester, Rami Al-Rfou, and Noah Constant.
\newblock The power of scale for parameter-efficient prompt tuning.
\newblock In {\em EMNLP}, 2021.

\bibitem[\protect\citeauthoryear{Levesque \bgroup \em et al.\egroup
  }{2012}]{Levesque2011TheWS}
Hector Levesque, Ernest Davis, and Leora Morgenstern.
\newblock The winograd schema challenge.
\newblock In {\em International Conference on the Principles of Knowledge
  Representation and Reasoning}, 2012.

\bibitem[\protect\citeauthoryear{Li and Liang}{2021}]{li-liang-2021-prefix}
Xiang~Lisa Li and Percy Liang.
\newblock Prefix-tuning: Optimizing continuous prompts for generation.
\newblock In {\em ACL}, 2021.

\bibitem[\protect\citeauthoryear{Lin}{2004}]{Lin2004ROUGEAP}
Chin-Yew Lin.
\newblock Rouge: A package for automatic evaluation of summaries.
\newblock In {\em ACL}, 2004.

\bibitem[\protect\citeauthoryear{Liu \bgroup \em et al.\egroup
  }{2021}]{Liu2021GPTUT}
Xiao Liu, Yanan Zheng, Zhengxiao Du, Ming Ding, Yujie Qian, Zhilin Yang, and
  Jie Tang.
\newblock Gpt understands, too.
\newblock {\em ArXiv}, abs/2103.10385, 2021.

\bibitem[\protect\citeauthoryear{Novikova \bgroup \em et al.\egroup
  }{2017}]{Novikova2017TheED}
Jekaterina Novikova, Ondrej Dusek, and Verena Rieser.
\newblock The e2e dataset: New challenges for end-to-end generation.
\newblock In {\em SIGDIAL}, 2017.

\bibitem[\protect\citeauthoryear{Papineni \bgroup \em et al.\egroup
  }{2002}]{Papineni2002BleuAM}
Kishore Papineni, Salim Roukos, Todd Ward, and Wei-Jing Zhu.
\newblock Bleu: a method for automatic evaluation of machine translation.
\newblock In {\em ACL}, 2002.

\bibitem[\protect\citeauthoryear{Radev \bgroup \em et al.\egroup
  }{2021}]{Radev2021DARTOS}
Dragomir Radev, Rui Zhang, Amrit Rau, Abhinand Sivaprasad, Chia-Hsuan Hsieh,
  Nazneen Rajani, Xiangru Tang, Aadit Vyas, Neha Verma, Pranav Krishna,
  Yangxiaokang Liu, Nadia Irwanto, Jessica Pan, Faiaz Rahman, Ahmad Zaidi,
  Murori Mutuma, Yasin Tarabar, Ankit Gupta, Tao Yu, Yi~Chern Tan, Xi~Victoria
  Lin, Caiming Xiong, and Richard Socher.
\newblock Dart: Open-domain structured data record to text generation.
\newblock In {\em NAACL-HLT}, 2021.

\bibitem[\protect\citeauthoryear{Radford \bgroup \em et al.\egroup
  }{2019}]{radford2019language}
Alec Radford, Jeffrey Wu, Rewon Child, David Luan, Dario Amodei, Ilya
  Sutskever, et~al.
\newblock Language models are unsupervised multitask learners.
\newblock {\em OpenAI blog}, 1(8), 2019.

\bibitem[\protect\citeauthoryear{Schick and
  Sch{\"u}tze}{2021a}]{schick-schutze-2021-exploiting}
Timo Schick and Hinrich Sch{\"u}tze.
\newblock Exploiting cloze-questions for few-shot text classification and
  natural language inference.
\newblock In {\em EACL}, 2021.

\bibitem[\protect\citeauthoryear{Schick and
  Sch{\"u}tze}{2021b}]{schick-schutze-2021-just}
Timo Schick and Hinrich Sch{\"u}tze.
\newblock It{'}s not just size that matters: Small language models are also
  few-shot learners.
\newblock In {\em NAACL-HLT}, 2021.

\bibitem[\protect\citeauthoryear{Shin \bgroup \em et al.\egroup
  }{2020}]{shin-etal-2020-autoprompt}
Taylor Shin, Yasaman Razeghi, Robert~L. Logan~IV, Eric Wallace, and Sameer
  Singh.
\newblock {A}uto{P}rompt: {E}liciting {K}nowledge from {L}anguage {M}odels with
  {A}utomatically {G}enerated {P}rompts.
\newblock In {\em EMNLP}, 2020.

\bibitem[\protect\citeauthoryear{Snover \bgroup \em et al.\egroup
  }{2006}]{snover-etal-2006-study}
Matthew Snover, Bonnie Dorr, Rich Schwartz, Linnea Micciulla, and John Makhoul.
\newblock A study of translation edit rate with targeted human annotation.
\newblock In {\em Proceedings of the 7th Conference of the Association for
  Machine Translation in the Americas: Technical Papers}, 2006.

\bibitem[\protect\citeauthoryear{Tam \bgroup \em et al.\egroup
  }{2021}]{tam-etal-2021-improving}
Derek Tam, Rakesh R.~Menon, Mohit Bansal, Shashank Srivastava, and Colin
  Raffel.
\newblock Improving and simplifying pattern exploiting training.
\newblock In {\em EMNLP}, 2021.

\bibitem[\protect\citeauthoryear{Trinh and Le}{2018}]{Trinh2018ASM}
Trieu~H. Trinh and Quoc~V. Le.
\newblock A simple method for commonsense reasoning.
\newblock {\em ArXiv}, abs/1806.02847, 2018.

\bibitem[\protect\citeauthoryear{Vedantam \bgroup \em et al.\egroup
  }{2015}]{Vedantam2015CIDErCI}
Ramakrishna Vedantam, C.~Lawrence Zitnick, and Devi Parikh.
\newblock Cider: Consensus-based image description evaluation.
\newblock In {\em CVPR}, 2015.

\bibitem[\protect\citeauthoryear{Wang \bgroup \em et al.\egroup
  }{2019}]{Wang2019SuperGLUEAS}
Alex Wang, Yada Pruksachatkun, Nikita Nangia, Amanpreet Singh, Julian Michael,
  Felix Hill, Omer Levy, and Samuel~R. Bowman.
\newblock Superglue: A stickier benchmark for general-purpose language
  understanding systems.
\newblock In {\em NeurIPS}, 2019.

\bibitem[\protect\citeauthoryear{Wolf \bgroup \em et al.\egroup
  }{2020}]{Wolf2020TransformersSN}
Thomas Wolf, Lysandre Debut, Victor Sanh, Julien Chaumond, Clement Delangue,
  Anthony Moi, Pierric Cistac, Tim Rault, R{\'e}mi Louf, Morgan Funtowicz, and
  Jamie Brew.
\newblock Transformers: State-of-the-art natural language processing.
\newblock In {\em EMNLP}, 2020.

\bibitem[\protect\citeauthoryear{Zhang \bgroup \em et al.\egroup
  }{2020}]{Zhang*2020BERTScore:}
Tianyi Zhang, Varsha Kishore, Felix Wu, Kilian~Q. Weinberger, and Yoav Artzi.
\newblock Bertscore: Evaluating text generation with bert.
\newblock In {\em ICLR}, 2020.

\bibitem[\protect\citeauthoryear{Zhong \bgroup \em et al.\egroup
  }{2021}]{zhong-etal-2021-factual}
Zexuan Zhong, Dan Friedman, and Danqi Chen.
\newblock Factual probing is [{MASK}]: Learning vs. learning to recall.
\newblock In {\em NAACL-HLT}, 2021.

\end{thebibliography}

\end{document}